\documentclass{ecai} 

\usepackage{latexsym}
\usepackage{amssymb}
\usepackage{amsmath}
\usepackage{amsthm}
\usepackage{booktabs}
\usepackage{enumitem}
\usepackage{graphicx}
\usepackage{color}

\usepackage{subcaption} 
\usepackage{caption} 
\usepackage{ulem}

\newcommand{\BibTeX}{B\kern-.05em{\sc i\kern-.025em b}\kern-.08em\TeX}

\begin{document}


\begin{frontmatter}


\paperid{123} 


\title{Fair Graph Neural Network with Supervised Contrastive Regularization}


\author[A]{\fnms{Mahdi}~\snm{Tavassoli Kejani}\orcid{0000-0002-7171-5746}\thanks{Corresponding Author. Email: mahdi.tavassoli-kejani@univ-toulouse.fr.}}
\author[B,C]{\fnms{Fadi}~\snm{Dornaika}\orcid{0000-0001-6581-9680
}} 
\author[A]{\fnms{Jean-Michel}~\snm{Loubes}\orcid{0000-0002-1252-2960}}

\address[A]{Institut de Mathématiques de Toulouse, Université Toulouse 3, Toulouse, France}
\address[B]{University of the Basque Country UPV/EHU, San Sebastian, Spain}
\address[C]{IKERBASQUE, Basque Foundation for Science, Bilbao, Spain}

\onecolumn

\begin{abstract}
In recent years, Graph Neural Networks (GNNs) have made significant advancements, particularly in tasks such as node classification, link prediction, and graph representation. However, challenges arise from biases that can be hidden not only in the node attributes but also in the connections between entities. Therefore, ensuring fairness in graph neural network learning has become a critical problem. To address this issue, we propose a novel model for training fairness-aware GNN, which enhances the Counterfactual Augmented Fair Graph Neural Network Framework (CAF). Our approach integrates Supervised Contrastive Loss and Environmental Loss to enhance both accuracy and fairness. Experimental validation on three real datasets demonstrates the superiority of our proposed model over CAF and several other existing graph-based learning methods.


\end{abstract}

\end{frontmatter}


\section{Modeling fairness using Graph Neural Networks} 
\subsection{A review of fairness and Graph Neural Networks}
Graph Neural Networks (GNNs) have emerged as a powerful paradigm for learning node representations on graphs, significantly advancing the state-of-the-art in tasks such as node classification, link prediction, and graph classification. By efficiently capturing the structural information within graphs, GNNs have found widespread application across various domains, including social network analysis, recommendation systems, drug discovery, and more.

Nowadays, Graph Neural Networks (GNNs)  are a class of deep learning models designed to operate on data represented in graph form. Graphs consist of nodes (or vertices) and edges connecting these nodes, making them ideal for representing complex relational and structural information. The research of GNN began in \cite{scarselli2008graph} for a review. They faced several drawbacks \cite{wu2020comprehensive, gupta2021graph} such as model depth , scalability, convergence and more, but their construction  have made significant progress over the years. However, challenges persist due to biases that may be embedded in node attributes or edges between nodes, resulting in unfair predictions with serious consequences. For example, in book recommendation with graph neutral network, where the GNN methods could be biased towards suggesting books with male authors \cite{buyl2020debayes}. Therefore, it is crucial to ensure that these models and resulting representations are safe and reliable. Addressing bias and promoting fairness in GNNs are thus vital research areas.

However, as GNNs leverage historical data to learn representations, there is a growing concern that they may inadvertently perpetuate and even amplify existing biases present in the training data. This issue of fairness in GNNs is particularly critical because graphs often represent social structures, interactions, and personal relationships, where biases towards certain groups or individuals can have profound implications. For instance, in a social network, a biased model could lead to unfair recommendations, while in financial services, it might result in inequitable credit scoring.

Fairness has been studied over the last years with several methods to obtain fair decisions.  Many methods try to balance the effect of unwanted correlations with variables that drive the bias in the sense that the decisions are different when the variable change while from moral, legal or ethical reasons, there should be no change. This variable is assumed to model a characteristic of the individuals that may convey  bias leading to possible discrimination as defined in \cite{chouldechova2020snapshot}, \cite{3495724.3497012}, \cite{barocas2017fairness} or \cite{besse2019can,besse2021survey} and references therein. In the literature on fairness in machine learning, such variable divides the individuals into a majority group corresponding to $S=1$ and a minority group $S=0$. The variable is often referred to as the sensitive attribute.  The influence of this variable on the learnt algorithm may induce unwanted changes in the decision which can be measured using a statistical parity measure, or different level of performance for different subgroups measured by equality of odds measures. \\
Many authors propose several ways to tackle this issue. We refer for instance to \cite{dwork2012fairness}, \cite{zafar2017fairness}, \cite{RisserEtAlJMIV2022}, \cite{gordaliza2019obtaining} and references therein, which propose several techniques to add fairness constraint when learning the algorithm. Another general way to achieve fairness is to use a latent space which disentangles the effect of the characteristics and the sensitive variables. This idea is developed for instance  in \cite{amini2019uncovering}, \cite{zemel2013learning}. Such ideas are very similar to the notions of counterfactual in  fairness as presented in \cite{kusner2017counterfactua}, \cite{de2021transport} where the authors propose to model how changes in the sensitive variable can be modeled as a causal intervention enabling to respond to the question {\it what-if} the sensitive variable had been different and what would have been the corresponding decision. This counterfactual definition of fairness implies the so-called statistical parity fairness and will be related to the notion of fairness for graph neural networks.

Recent research has begun to address these concerns by investigating the sources of bias in graph data and developing methodologies to mitigate discriminatory biases in GNN-based applications. A notable direction in this field includes the identification and quantification of bias in node embeddings, exploring how structural properties of graphs, such as homophily and community structure, can contribute to bias amplification. Other publications offer insights on how to address fairness in the construction of the GNNs. For instance, \cite{venkatasubramanian2021fairness, kang2021fair, kang2022algorithmic} introduced the concept of fairness in graph representation learning, highlighting the need for fairness-aware algorithms in GNNs. Furthermore, research by \cite{khajehnejad2022crosswalk} proposed a framework for fair node classification, which adapts adversarial debiasing techniques to the graph setting, demonstrating the potential for achieving fairness in GNN predictions. Optimal transport methods popular in fairness as in \cite{gordaliza2019obtaining} have been successfully applied for graph and edges prediction tasks in \cite{laclau2021all}.\vskip .1in

Efforts to mitigate bias in graph-structured data have led to the development of fairness-aware GNNs \cite{dai2022comprehensive, guo2023counterfactual, wu2020comprehensive}. For instance, FairGNN \cite{dai2021say} employs adversarial learning to ensure node representations remain independent of sensitive attributes. Another notable approach is EDITS \cite{dong2022edits}, which concentrates on addressing bias in both the adjacency matrix and data matrix as part of the pre-processing stage. Additionally, Fairwalk \cite{rahman2019fairwalk} utilizes a fair random walk method to learn fair network embeddings.

Recent studies have turned their attention to attaining counterfactual fairness within graph data. In this context, a classification prediction for an individual is considered counterfactually fair when it stays consistent across both the actual and counterfactual worlds. For instance, NIFTY \cite{agarwal2021towards} maximizing the similarity between representations of the original nodes in the graph and their counterparts in the augmented graph. These counterparts are generated by either slightly perturbing the original node attributes and edges or considering counterfactuals of the original nodes where the value of the sensitive attribute is modified and GEAR \cite{Ma_2022} adopts GraphVAE \cite{kipf2016variational} to generate counterfactuals corresponding to perturbations on each node's and their neighbors' sensitive attributes. It then enforces fairness by minimizing the discrepancy between the representations learned from the original graph and the counterfactuals for each node.

These approaches \cite{agarwal2021towards, kipf2016variational} can yield counterfactuals that are potentially unrealistic, which, in turn, have the capacity to disrupt the underlying semantic structure and decrease the performance of the model. However, finding realistic counterfactuals presents several challenges: (i) Graph data is quite complex, making it infeasible to directly identify counterfactuals in the original data space. Additionally, some guidance or rules are needed to facilitate the discovery of counterfactuals. (ii) Achieving graph counterfactual fairness requires learned representations to be invariant to sensitive attributes and information causally influenced by sensitive attributes.

In CAF \cite{Guo_2023}, the authors address the aforementioned challenges by constructing a Structural Causal Model to identify counterfactuals and learn disentangled representations. They divided these representations into content and environmental information parts, ensuring that the content information remains informative for labels and invariant to sensitive attributes, while the environmental information depends on sensitive attributes. Despite constructing a fair model based on these principles using realistic counterfactuals, experimental results from CAF indicate that while fairness is improved, performance is decreased. \vskip .1in

The results presented in the CAF paper indicate that while improvements were made in fairness, the price for achieving fair decisions leads to a drastic decrease of the accuracy of the decisions This motivated us to extend this model by introducing two extra constraints aimed at simultaneously improving fairness and accuracy, leading to a better balance preserving the efficiency of the algorithm. Our main contributions are:
\begin{enumerate}
    \item We use the Supervised Contrastive Loss to encourage similar content information with similar labels to have similar representations, irrespective of sensitive attributes.
    \item We use the Environmental Loss to encourage dissimilarity in environment information when the sensitive attributes are dissimilar.
    \item We propose a novel framework by extending the CAF framework and utilizing the Supervised Contrastive Loss and Environmental Loss. We demonstrate the effectiveness of our model through experiments on real-world datasets and comparison with other methods.
\end{enumerate}
\subsection{Notations}
In this section, we present the notations employed in this paper.  We assume we observe $i=1,\dots,n$ individuals with characterizes $x_i \in \mathbb{R}^d$ for $d>0$.  Let define the matrix of observations $\mathcal{X}$  formed by combining vectors $[x_1, x_2, \ldots, x_n]$ in $\mathbb{R}^{n \times d}$. The aim of our study is to forecast a binary label $Y \in \{0,1\}$ which is only observed for some individuals. In addition, the individuals  are also characterized by an observed binary variable  $S \in \{0, 1\}$. The influence of the sensitive variable on the predictions is at the heart of our study. Our aim is to mitigate its effect and build a latent space where the effects of the features and the sensitive variables are separated, enabling bias control and bias mitigation. \\
We will model the relationships between individuals using a graph where the nodes are the observed individuals.  A graph is defined as a couple $G = (\mathcal{V}, \mathcal{E})$, where $\mathcal{V}$ composed by a set of nodes of size $|\mathcal{V}|=n$,  $\mathcal{E}$ and a  set of edges. The nodes are connected by edges and the connections are modeled using a matrix $\mathcal{A} \in \mathbb{R}^{n \times n}$ known as  the adjacency matrix of the graph $G$. As usual  $\mathcal{A}_{i,j} = 1$ if edge $i \rightarrow j$ exists, otherwise $\mathcal{A}_{i,j} = 0$. Further we assume that $G$ is undirected and unweighted. We refer for instance to \cite{bondy2008graph} for definitions of graphs. The individuals $i=1,\dots,n$ are characterized by their features denoted $x_i \in \mathbb{R}^d$ that can be used or not in the construction of the initial graph. \\
 We are in a semi-supervised setting where $\{1,\dots,n\}=\mathcal{L} \cup \mathcal{U}$, with $\mathcal{L}=\{1,\dots,l\}$ contains the indices of the labeled samples while $\mathcal{U}=\{l+1,\dots,l+u\}$ is the set of indices of the unlabeled samples. \\
In this paper we only present the case of a binary sensitive attribute . Yet, following the works \cite{Ma_2022} and \cite{Guo_2023} our model seamlessly extends to handle various types of weighted graphs and sensitive attributes, not only the binary case.
\section{Constrastive Loss for Fair Graph Neural Networks}
\subsection{Setting a Contrastive Graph Neural Network}
Traditional methods for learning node representations utilize an encoder $f_{\theta}(\cdot): \mathcal{X}^{n \times d} \times \mathcal{A}^{n \times n} \to \mathcal{H}^{n \times d'}$ to map each node to a latent lower dimensional  representation so that it preserves both node attributes and local graph structure information. Here, $f_{\theta}$ represents graph neural network (GNN) architectures, such as GCN \cite{kipf2016semi} and GraphSAGE \cite{hamilton2017inductive}, and $\mathcal{H}$ denotes the latent representations.

In the study by Guo et al. \cite{Guo_2023}, their objective is to ensure  that label predictions are independent from   chosen sensitive attributes, hence obtaining so-called {\it fair} prediction. For this, they introduce the Counterfactual Augmented Fair Graph Neural Network Framework (CAF) for learning fair node representations. Fairness is  achieved by imposing several constraints and mainly partitioning the latent representation $\mathcal{H}$ into two distinct components: $C$ (content information) and $E$ (environmental information). This partition of $\mathcal{H}$ is formally represented as follows:

\begin{eqnarray}
    f_{\theta}(\mathcal{X}, \mathcal{A}) = \mathcal{H} = [ {C} , {E} ]
\end{eqnarray}

The first $d_{c}$ columns of $H$ constitute the content information matrix $C$, which contains information relevant for label prediction while remaining unaffected by changes of sensitive attributes. This  content information ($C$) is the only one that will be used  for prediction. Actually a classifier $f_{\phi}$ parameterized by $\phi$ is learnt by using the content information as input while forecasting the class distribution of each node as follows:

\begin{eqnarray} \label{eq:forecast}
     f_{\phi}(c) = \hat{y}
\end{eqnarray}

While the next $d_{e}$ columns form the environment information matrix $E$ which is designed to capture environmental information primarily dependent on sensitive attributes. Notably, $d_c$ is equal to $d_e$.\\

\begin{figure*}[h]
    \centering
    \includegraphics[width=0.8\textwidth]{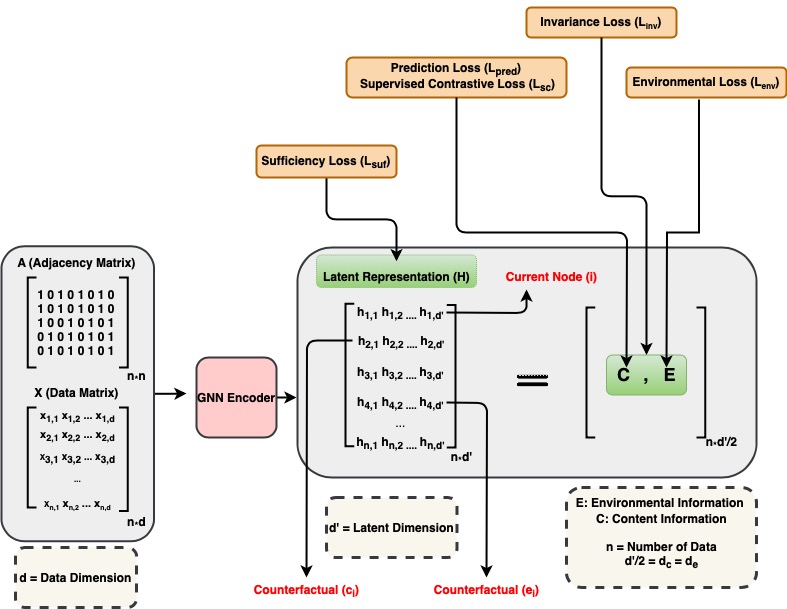}
    \caption{An illustration of our proposed framework}
    \label{fig:model}
\end{figure*}

For a given nodes $i$ we can consider two notions of counterfactual nodes: we can either look  at the closest nodes with similar features and the same outputs of the algorithms but with distinct sensitive attributes  or  the closest nodes with similar features, the same sensitive attributes but different outputs of the algorithm. We want to design the embedding in a such a way that, in the first case, the changes in the features $h$ will mainly be changes with respect to environment part $e$ since this part should be related to the sensitive attribute. On the contrary, we expect, in the second case, that the changes will be driven by the $c$ part of the features, i.e the content which explain the decision. Hence we will use the following notations 
\begin{eqnarray}
h_{i}^{e} \in \text{argmin}_{h_j \in H} \{\|h_i - h_j\|_2^2 \,|\, \hat{y}_i = \hat{y}_j, \, s_i \neq s_j\}
\end{eqnarray}
\begin{eqnarray}
h_{i}^{c} \in \text{argmin}_{h_j \in H} \{\|h_i - h_j\|_2^2 \,|\, \hat{y}_i \neq \hat{y}_j, \, s_i = s_j\}
\end{eqnarray}
and we note $h_{i}^{e}=(c_i^e,e_i^e)$ and respectively $h_{i}^{c}=(c_i^c,e_i^c)$.

 We point out that the corresponding counterfactuals are not uniquely defined for each node. We denote by $\mathcal{K}^e(i)$ and $\mathcal{K}^c(i)$ the sets of counterfactuals of a node $i$ and we set   ${K}$ the same number of chosen counterfactuals. \vskip .1in
We will build a Contrastive Counterfactually Fair Graph Neural Network by considering the following losses.
The first term $\mathcal{L}_{\text{pred}}$ is the usual Cross Entropy Loss, referred to as the prediction loss. \\
\begin{eqnarray}
    \mathcal{L}_{\text{pred}} = \frac{1}{\lvert \mathcal{L} \rvert} \sum_{i \in \mathcal{L}} \ell(\hat{y}_i, y_i).
\end{eqnarray}
 Note also that computing the counterfactuals requires knowing the estimation of the labels for all points. Hence,  due to the semi-supervised setting we have a limited number of labels in the training set for counterfactual guidance,  and thus the model is pre-trained to obtain pseudo-labels $\hat{y}$ by applying the following loss function to Equation~(\ref{eq:forecast}). \\
 
To learn fair latent representations the loss is designed to achieve the decomposition into content and environment variables. Namely the purpose is that  when the sensitive attribute $s_i$ of a node $i$ is flipped to $1 - s_i$, its content representation $c_i$ is expected to remain constant, while its environment representation $e_i$ changes correspondingly, resulting in the formation of the counterfactual $h_{i}^{e}$. Similarly, when the label $y_i$ of a node $i$ is flipped while keeping the sensitive attribute $s_i$ unchanged, the content representation $c_i$ is changed, while the environment representation $e_i$ remain constant, forming $h_{i}^{c}$.

The second term, $\mathcal{L}_{\text{inv}}$, is the invariance loss, defined as follows:
\begin{equation*}
    \mathcal{L}_{\text{inv}} = \frac{1}{|\mathcal{V}|\cdot K}  \sum_{v_i \in V} \sum_{k=1}^{K} \left[ \text{dis}(c_i, c_{i}^{e_{k}}) 
+  \text{dis}(e_i, e_{i}^{c_{k}}) + \gamma K\cdot \left| \cos(c_i, e_i) \right| \right]
\end{equation*}
where $\text{dis}(\cdot, \cdot)$ is a distance metric, such as the cosine distance and $L2$ distance. $|\cos(\cdot, \cdot)|$ represents the absolute value of the cosine similarity, which imposes an orthogonality constraint, and $\gamma$ is the hyper-parameter that controls it. This loss promotes fairness of the embedding.

The third term, $\mathcal{L}_{\text{suf}}$, the sufficiency loss, which is defined in order to preserve the graph information and ensure that the latent representation has the same topological property as the initial  input graph. It is defined as follows:
\begin{equation*}
    \mathcal{L}_{\text{suf}} = \frac{1}{|\mathcal{A}| + |\mathcal{A}^-|} 
    \times \sum_{({i}, {j})\in \mathcal{A} \cup \mathcal{A}^-} -a_{i,j} \log p_{i,j} - (1 - a_{i,j}) \log (1 - p_{i,j})
\end{equation*}
where $\mathcal{A}^-$ is the set of sampled negative edges. If nodes $i$ and $j$ are connected, $a_{i,j} = 1$; otherwise, $a_{i,j} = 0$. Additionally, $p_{i,j} = \sigma(h_i h_j^T)$ represents the link existence probability, where $\sigma$ denotes the Sigmoid function.

Note that Guo et al. \cite{Guo_2023}, uses the following loss function to learn fair classification 
\begin{eqnarray}
\min_{\theta, \phi} \mathcal{L} = \mathcal{L}_{\text{pred}} + \alpha \mathcal{L}_{\text{inv}} + \beta \mathcal{L}_{\text{suf}}
\end{eqnarray}
where $\theta$ and $\phi$ are GNN encoder and prediction head parameters, and $\alpha$ and $\beta$ are hyper-parameters controlling invariance and sufficiency.   Although CAF can enhance fairness, it tends to reduce accuracy since it ocuses solely on splitting content and environment information, ensures that the sensitive attribute does not impact node classification tasks.
\subsection{Improvements using Contrastive Loss functions}
In our study we go on improving their model by incorporating two losses that will be referred to as the Supervised Contrastive Loss and the Environmental Loss and prove that they enable to mitigate  fairness without harming accuracy. Our architecture is displayed in Figure ~(\ref{fig:model}).

Supervised Contrastive Loss, $\mathcal{L}_{\text{sc}}$, is a crucial element in Supervised Contrastive Learning \cite{khosla2020supervised}. It works by encouraging similar data points to have similar representation while pushing dissimilar data point apart in the Latent space. Using this property, this loss has been successfully used to model counterfactuals since it encourages separation of the points in the embedding space based on the choice of a relevant similarity. For instance in \cite{todo2023counterfactual}, it enables to represent time series sharing into an embedding space made of two components, one providing a common pattern while the other models the difference between the functions. In the context of graph representation,  we apply this loss specifically to the content information (C) of a latent representation, defines as follows 
\begin{eqnarray}
    \mathcal{L}_{\text{sc}} = \sum_{i =1}^n \frac{-1}{\lvert \mathcal{P}(i) \rvert} \sum_{p \in \mathcal{P}(i)} \log\frac{\exp\left( \frac{\mathbf{c}_i \cdot \mathbf{c}_p}{\tau} \right)}{\sum_{a \in \mathcal{V}(i)} \exp\left( \frac{\mathbf{c}_i \cdot \mathbf{c}_a}{\tau} \right)}
\end{eqnarray}
where $\mathcal{P}(i) \equiv \{ p \in \mathcal{V}(i): \hat{\mathbf{y}}_{p} \equiv \hat{\mathbf{y}}_{i} \}$ is the set of indices of all positive samples (those with the class labels).  Nodes sharing similar content and similar labels tend to have similar embedding in the latent space.

In our proposed model, we apply this loss during the pretrained phase using labeled data. However, during the training phase, we apply this loss to all training data, utilizing pseudo-labels generated during the pretrained process.

The Environmental Loss, $\mathcal{L}_{\text{env}}$, works by encouraging node has dissimilar sensitive attribute to have dissimilar environment information in the latent space. $\mathcal{L}_{\text{env}}$ is defined as follows:
\begin{eqnarray}
    \mathcal{L}_{\text{env}} =  \frac{1}{n} \sum_{i=1}^n \frac{1}{K'} \sum_{j=1}^{K'} \text{dis}(e_i, e_j)
\end{eqnarray}
Where $K^{\prime}$ denotes the number of nodes that have dissimilar sensitive attributes but whose latent representations are close to the current node.

In our proposed model, we apply this loss during the pretrained and training phase.

According to the above definition, the loss function of our proposed model for the pretraining phase is defined as follows:
\begin{eqnarray}
\min_{\theta, \phi} \mathcal{L} = \mathcal{L}_{\text{pred}} + \omega \mathcal{L}_{\text{sc}} - \eta \mathcal{L}_{\text{env}}
\end{eqnarray}
and for the training phase is :
\begin{eqnarray}
\min_{\theta, \phi} \mathcal{L} = \mathcal{L}_{\text{pred}} + \alpha \mathcal{L}_{\text{inv}} + \beta \mathcal{L}_{\text{suf}} + \omega \mathcal{L}_{\text{sc}} - \eta \mathcal{L}_{\text{env}}
\end{eqnarray}
All the parameters $\omega$, $\eta$, $\alpha$, $\beta$, $\gamma$ are trade-off parameters that balance the influence of all terms. \\
Their influence and the choice is described in the following section. \vskip .1in

As mentioned, we divide the latent representation into content and environmental information parts. Fairness and performance improve when the content information contains no sensitive attribute-related information, and all such information exists in the environmental part. Therefore, we utilize the Supervised Contrastive Loss and Environmental Loss to enforce that each part has related information.
\section{Experiments}
\begin{table*}[!ht]
    \centering
    \caption{The evaluation of node classification and group fairness performance across real-world datasets}
    \label{TABLE_RESULTS}
    \resizebox{\textwidth}{!}
    {
        \begin{tabular}{l rrrrrrrrr}
            \hline
            \textbf{\footnotesize Dataset \hspace{-0.1mm}\textbackslash \hspace{-0.1mm} Method} & \textbf{Metric} & \textbf{GCN} & \textbf{GraphSAGE} & \textbf{GIN} & \textbf{FairGNN} & \textbf{EDITS} & \textbf{GEAR} & \textbf{CAF} & \textbf{SCCAF} \\
            \hline
            {\textbf{GERMAN}} & AUC ($\uparrow$) & 57.38 $\pm$ 2.85  & \uline{62.14 $\pm$ 5.35} & 59.16 $\pm$ 5.98 & 54.28 $\pm$ 3.99 & 50.75 $\pm$ 4.06 & 44.35 $\pm$ 5.88 & 59.76 $\pm$ 5.88 & \textbf{66.07 $\pm$ 2.97} \\
            & F1 ($\uparrow$) & 65.51 $\pm$ 1.45 & 73.15 $\pm$ 14.18 & 72.40 $\pm$ 3.91 & 62.40 $\pm$ 9.14 & 62.34 $\pm$ 6.09 & 50.78 $\pm$ 21.61 & \uline{81.38 $\pm$ 1.22} & \textbf{82.38 $\pm$ 0.11} \\
            & $\Delta_{SP} (\downarrow)$ & 5.13 $\pm$ 4.30 & 4.09 $\pm$ 4.56 & 6.82 $\pm$ 1.98 & 8.76 $\pm$ 7.82 & 7.99 $\pm$ 5.88 & 8.38 $\pm$ 4.93 & \uline{3.93 $\pm$ 3.05} & \textbf{2.4 $\pm$ 3.11} \\
            & $\Delta_{EO} (\downarrow)$ & 5.71 $\pm$ 5.85 & 4.62 $\pm$ 4.57 & \uline{7.03 $\pm$ 3.94} & 10.44 $\pm$ 8.73 & 6.92 $\pm$ 6.02 & 8.29 $\pm$ 4.37 & \uline{3.73 $\pm$ 1.05} &  \textbf{1.1 $\pm$ 1.34} \\
            \hline
            {\textbf{BAIL}} & AUC ($\uparrow$) & 93.69 $\pm$ 1.27 & 97.17 $\pm$ 0.87 & 94.77 $\pm$ 0.82 & 94.87 $\pm$ 1.07 & 92.36 $\pm$ 1.40 & 86.86 $\pm$ 1.65 & \uline{97.45 $\pm$ 0.09} & \textbf{98.01 $\pm$ 0.49} \\
            & F1 ($\uparrow$) & 84.57 $\pm$ 1.34 & 93.40 $\pm$ 1.60 & 89.06 $\pm$ 1.67 & 87.57 $\pm$ 2.67 & 82.75 $\pm$ 2.43 & 79.00 $\pm$ 2.47 & \uline{94.43 $\pm$ 0.63} & \textbf{95.48 $\pm$ 1.88} \\
            & $\Delta_{SP} (\downarrow)$ & 6.88 $\pm$ 1.56 & 7.08 $\pm$ 1.17 & \uline{7.14 $\pm$ 1.59} & 7.40 $\pm$ 0.76 & \textbf{7.31 $\pm$ 1.15} & 7.95 $\pm$ 1.55 & 6.45 $\pm$ 1.48 & 7.54 $\pm$ 1.09 \\
            & $\Delta_{EO} (\downarrow)$ & 1.48 $\pm$ 0.93 & 1,14 $\pm$ 1.19 & 1.54 $\pm$ 1.26 & \uline{1.12 $\pm$ 0.76} & 1.94 $\pm$ 1.35 & 3,16 $\pm$ 1.58 & \uline{0.87 $\pm$ 1.21} & \textbf{0.64 $\pm$ 0.84} \\
            \hline
            {\textbf{CREDIT}} & AUC ($\uparrow$) & 67.36 $\pm$ 2.31 & \textbf{71.92  $\pm$ 1.95}  & 67.956 $\pm$ 6.43  & 63.19 $\pm$ 3.67 & 66.64 $\pm$ 0.35  &  69.17 $\pm$ 0.52  &  65.75 $\pm$ 2.38 &  \uline {70.03 $\pm$ 1.39} \\
            & F1 ($\uparrow$) & 80.06 $\pm$ 2.28 & 78.64 $\pm$ 0.71 & 72.28 $\pm$ 6.65 & 74.62 $\pm$ 1.07 & 84.42 $\pm$ 3.02 & 78.84  $\pm$ 1.02 & \uline{86.36$\pm$ 0.26} & \textbf{86.83 $\pm$ 0.15} \\
            & $\Delta_{SP} (\downarrow)$ & 12.93 $\pm$ 5.36 & 10.54 $\pm$ 4.42  &  8.79 $\pm$ 6.86 & 9.39 $\pm$ 2.65  & \textbf{ 3.16 $\pm$ 1.61 } &  14.298 $\pm$ 1.41 & \uline{3.43 $\pm$ 1.67} & 3.47 $\pm$ 2.22 \\
            & $\Delta_{EO} (\downarrow)$ &  11.40 $\pm$ 5.92 & 9.46 $\pm$ 4.36 & 8.87 $\pm$ 7.50 &  8.85 $\pm$ 3.12  & \uline{1.97 $\pm$ 1.43} & 13.24 $\pm$ 2.65 & 1.67 $\pm$ 1.93 & \textbf{1.52 $\pm$ 1.84} \\
        \end{tabular}
    }
\end{table*}
In this section, we assess the proposed model's performance on diverse tabular real-world datasets, encompassing German credit \cite{asuncion2007uci}, Bail \cite{jordan2015effect} and Credit Defaulter \cite{yeh2009comparisons}.
\subsection{Datasets}
\hspace*{1em} \textit{We used the following datasets:}
\begin{enumerate}
    \item \textbf{German Credit \cite{asuncion2007uci}:} 
          The dataset consists of 1000 individual clients from a German bank, and these individuals, acting as nodes, are linked when there is a significant similarity in their credit accounts. The main objective is to evaluate the credit risk level for each individual, emphasizing the sensitive attribute of 'gender'.
    \item \textbf{Bail \cite{jordan2015effect}:} 
          The dataset comprises 18,876 individuals granted bail by US state courts between 1990 and 2009. These individuals, serving as nodes, are linked based similarities in their demographics and prior criminal histories. The primary goal is to categorize defendants as either on bail or not, with a focus on the sensitive attribute 'race'.
    \item \textbf{Credit Defaulter \cite{yeh2009comparisons}:}
    The dataset includes 30,000 individual credit card holders, where these individuals, acting as nodes, are linked based on high similarity in payment information. The main objective is to predict whether an individual will default on credit card payments, with particular attention to the sensitive attribute of 'age'.
\end{enumerate}
\subsection{Experimental setup}
We compare the proposed SCCAF framework with state-of-the-art plain node classification methods, including GCN \cite{kipf2016semi}, GraphSAGE \cite{hamilton2017inductive}, and GIN \cite{xu2019powerful}. Additionally, we assess its performance against fair node classification methods (FairGNN \cite{dai2021say}, EDITS \cite{dong2022edits}), as well as graph counterfactual fairness method (GEAR \cite{ma2022learning}) and fair node representations for graph counterfactual fairness method (CAF \cite{Guo_2023})

For all databases, we adhere to the train/validation/test split methodology described in \cite{Guo_2023}. However, to ensure a more robust assessment of our method alongside others, we compute the average results obtained from five random splits of the data into train/validation/test sets. 

Each method under comparison is associated with its own set of hyperparameters, which undergo optimization through a grid search. Subsequently, the model demonstrating the best performance for each method is selected. This standardized approach is consistently applied across all methods being compared. Additionally, the learning rate is fixed at 0.01, and the random seed is kept constant.

For our proposed method, the parameters are set as follows: $K$ is selected from \{4, 6, 10\}, $K'$ is selected from \{10, 20\}, $\alpha$ is selected from \{0.2, 0.1, 0.07\}, $\beta$ is selected from \{0.1, 0.05, 0.01\}, $\gamma$ is set to 0.01, $\omega$ is set to 0.01, and $\eta$ is selected from \{0.1, 0.03\}

Moreover, to ensure impartial evaluations, all methods undergo testing using identical train/validation/test splits. This uniform methodology ensures fair and equitable comparisons across all evaluated methodologies.

\subsection{Related fairness metrics}
Researchers assess fairness for Graph Neural Networks across various fundamental levels. In their work \cite{chen2023fairness}, the authors classify fairness evaluation metrics for Graph Neural Networks into prediction-level metrics, graph-level metrics, neighborhood-level metrics and embedding-level metrics. In our work, we focus on prediction-level metrics, which we explain in this section.

Statistical Parity (SP) \cite{dai2021say}: The statistical parity (SP) metric, also known as demographic parity (DP), requires that predictions made by the model are independent of the sensitive attribute \( s \). It is defined as follows:
\begin{eqnarray}
    \Delta_{SP} = {\lvert\mathcal{P}({\hat{y}_{u} = 1 \lvert \ s = 0)}) - \mathcal{P}({\hat{y}_{u} = 1 \lvert \ s = 1)}\lvert}
\end{eqnarray}

Where \( \Delta_{SP} = 0 \) implies that all groups have the same selection rates, indicating complete fairness. Statistical parity measures the gap in preferential treatment between different groups. However, it's important to note that \( \Delta_{SP} \) does not consider whether individuals are qualified or not, as it does not take into account the ground-truth label \( Y \).

Equal Opportunity (EQ) \cite{dai2021say}: This principle dictates that the probability of a positive outcome being predicted for instances in a positive class should be the same for all subgroup members. It is defined as follows:
\begin{equation*}
    \Delta_{EO} = \left\lvert \mathcal{P}(\hat{y} = 1 \,|\, {y} = 1 , s = 0)  -  \mathcal{P}(\hat{y} = 1 \,|\, {y} = 1 , s = 1) \right\rvert
\end{equation*}

Where \( \Delta_{EQ} = 0 \) implies equal true positive rates across subgroups, indicating fairness.

\subsection{Method comparison}
We assess the model's performance. Table (\ref{TABLE_RESULTS}) presents the AUC and F1 score as percentages, along with their standard deviations across five random splits, providing insights into node classification proficiency. In terms of fairness evaluation, guided by \cite{10.1145/3437963.3441752}, we employ two widely recognized group fairness metrics specifically, statistical parity (SP) $\Delta_{SP}$ and equality parity (EO) $\Delta_{EO}$. 

\begin{figure*}[ht]
    \centering
    \begin{minipage}[b]{0.3\textwidth}
        \centering
        \begin{subfigure}[b]{\linewidth}
            \centering
            \includegraphics[width=\linewidth]{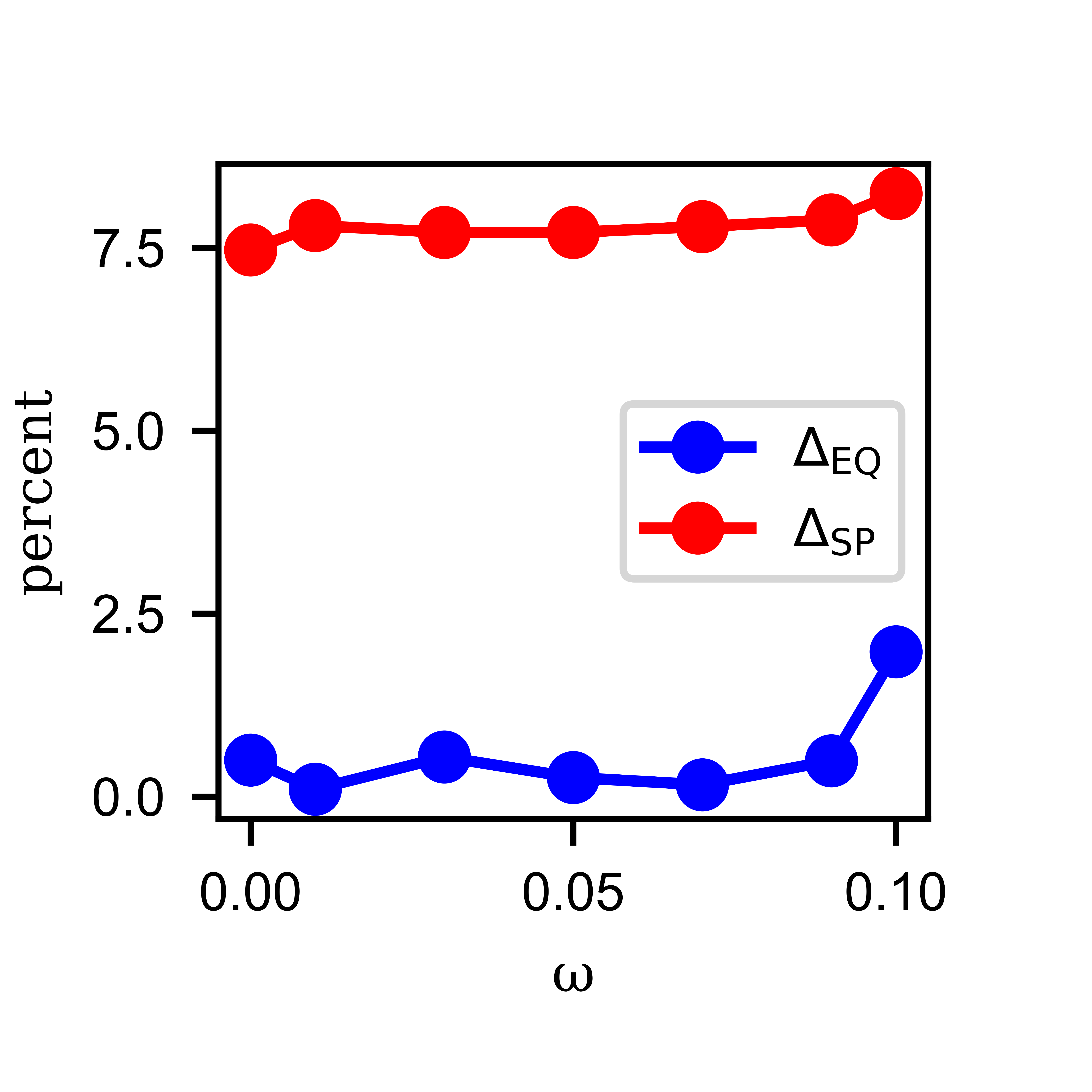}
            \label{fig:credit_parity}
        \end{subfigure}
        \begin{subfigure}[b]{\linewidth}
            \centering
            \includegraphics[width=\linewidth]{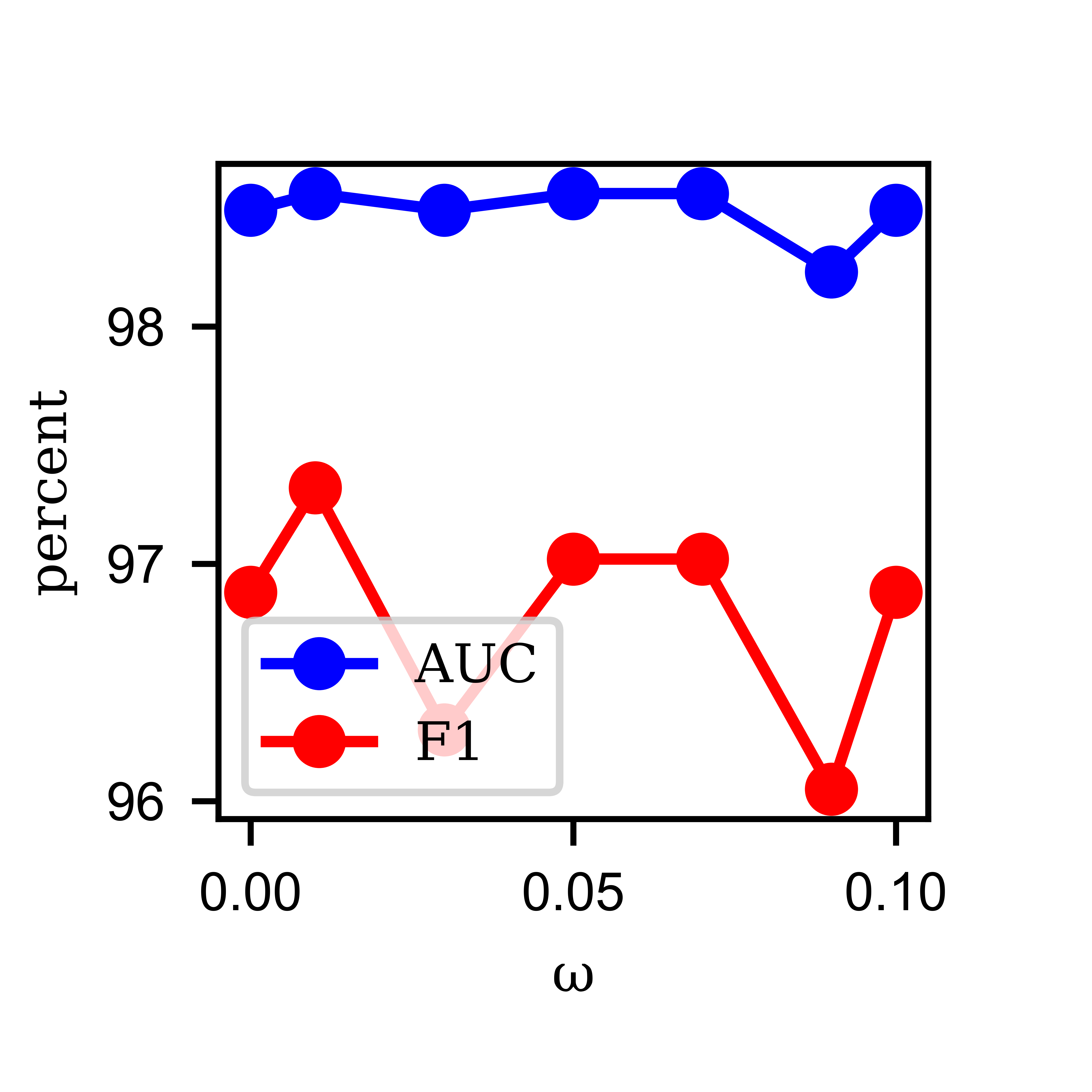}
            \label{fig:credit_accuracy}
        \end{subfigure}
    \end{minipage}
    \begin{minipage}[b]{0.3\textwidth}
        \centering
        \begin{subfigure}[b]{\linewidth}
            \centering
            \includegraphics[width=\linewidth]{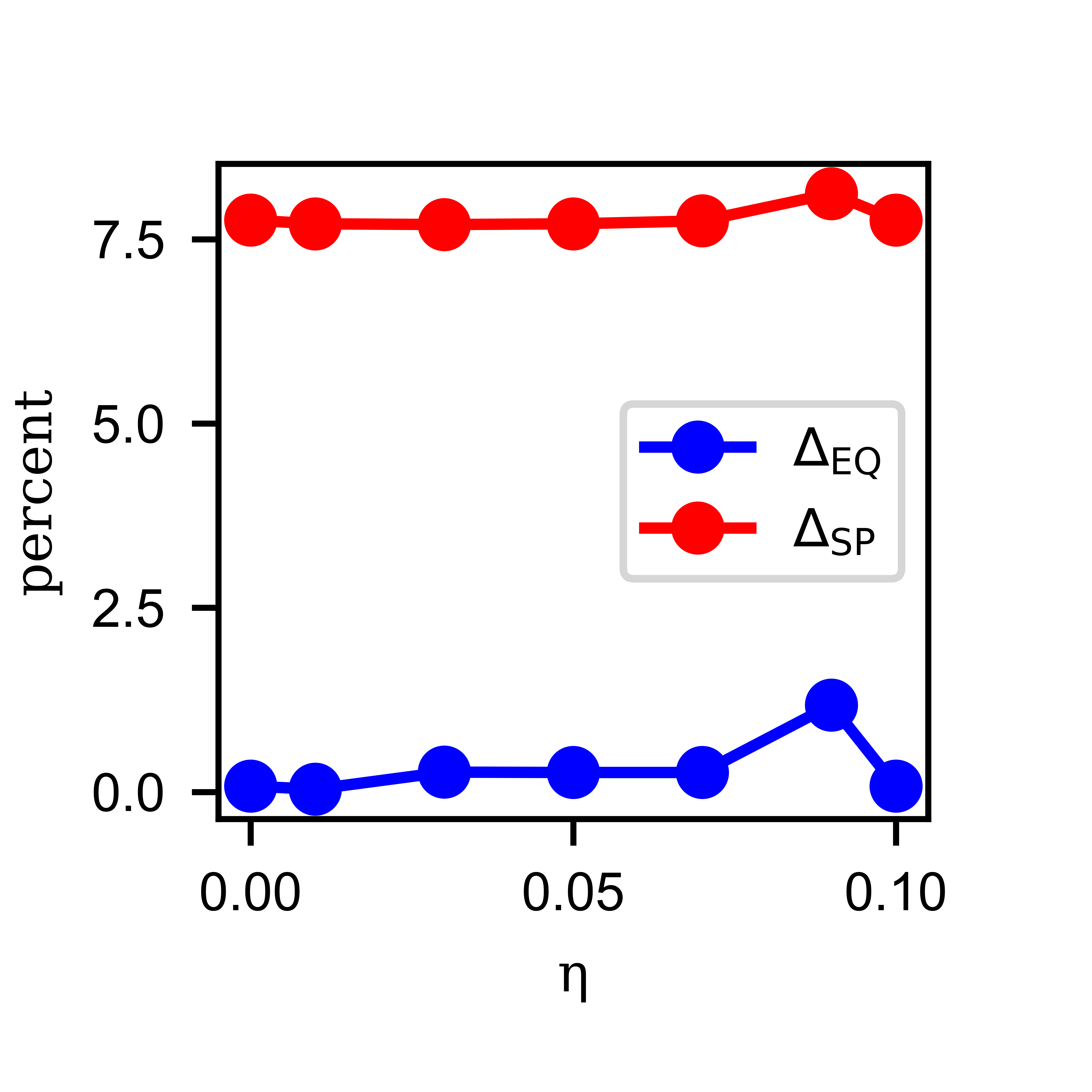}
            \label{fig:bail_parity}
        \end{subfigure}
        \begin{subfigure}[b]{\linewidth}
            \centering
            \includegraphics[width=\linewidth]{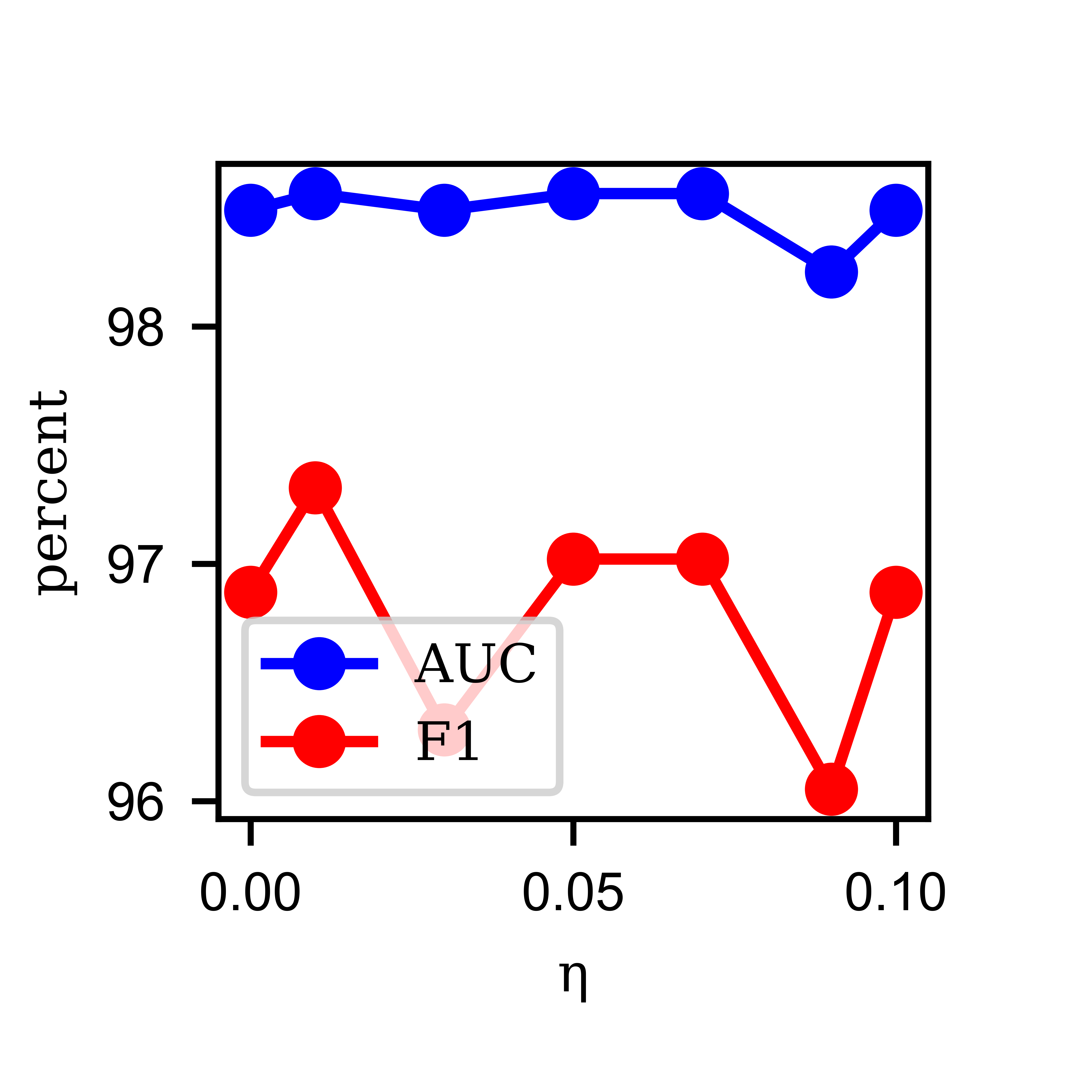}
            \label{fig:bail_accuracy}
        \end{subfigure}
    \end{minipage}
    \caption{$\Delta_{EQ}$, $\Delta_{SP}$, AUC, and F1 percentages of the proposed model SCCAF as functions of the parameter $\omega$ and $\eta$ in the Bail databases}
    \label{fig:comparison}
\end{figure*}

\subsection{Effect of the proposed SCCAF for fair classification task}
The results in Table (\ref{TABLE_RESULTS}) demonstrate that our proposed method can improve performance and fairness simultaneously. In the German database, we enhance the AUC and F1 metrics, related to performance, from 59.76 and 81.38 in the CAF method to 66.07 and 82.38, respectively, while $\Delta_{SP}$ and $\Delta_{EQ}$, related to fairness metrics, decrease from 3.93 and 3.73 to 2.4 and 1.1. In the Bail database, we enhance the AUC and F1 metrics, related to performance, from 97.45 and 94.43 in the CAF method to 98.01 and 95.48, respectively, and $\Delta_{EQ}$ metric decreases from 0.87 to 0.64. In the Credit database, we enhance the AUC and F1 metrics, related to performance, from 65.75 and 86.36 in the CAF method to 70.03 and 86.83, respectively, and $\Delta_{EQ}$ metric decreases from 1.67 to 1.52. We can observe a significant improvement over other methods. Therefore, we can conclude that if we can separate the content and environmental information in the best ways, both performance and fairness increase simultaneously.

\subsection{Effect of the Supervised Contrastive Loss}
The proposed method incorporates two parameters: $\omega$, which determines the weight of the Supervised Contrastive Loss in the total loss, and $\eta$, which determines the weight of the Environmental Loss in the total loss. Figure ~(\ref{fig:comparison}) illustrates the accuracy and fairness performance as a function of $\omega$ while other parameters are fixed (with $\alpha = 0.2$, $\beta = 0.1$, $\gamma = 0.01$, and $\eta = 0.1$). Similarly, it shows the accuracy and fairness performance as a function of $\eta$ while other parameters are fixed (with $\alpha = 0.2$, $\beta = 0.1$, $\gamma = 0.01$, and $\eta = 0.1$), using the test data from the Bail database. Increasing $\omega$ and $\eta$ generally leads to a trade-off between prediction performance and fairness, although the changes are not as significant as those observed in CAF. Therefore, this paper aims to achieve our goal of improving accuracy and fairness simultaneously.

\section{Conclusion}
In this paper, we introduced a novel framework called SCCAF, which enhances the CAF model for learning fair node representations. Similarly to CAF model, the  main objective is still to achieve fairness by imposing a notion of distributional independence in the latent space. This is usually achieved by partitioning the latent representation into two parts : a content and an environmental information part. The content information contains no sensitive attribute-related data, while all such information resides in the environmental part. The crucial aspect lies in how we define the loss function to ensure both separation of the information and yet effective learning of both the content and environmental information parts with relevant data. The loss function defined in the CAF method is not sufficient leading to separation but poor accuracy due to a too large loss of information. \\
\indent Hence we proposed to first use a Supervised Contrastive Loss, which yet requires to estimate  the pseudo-labels of all samples. This approach improves the learning of the content information part. Furthermore, to improve the relevance of the environmental latent space and its connection with the sensitive variable, we incorporated an Environmental Loss to ensure that for different sensitive attributes, their environmental part also differs. By adding these losses to the CAF loss, we achieved our goal effectively, leading to increased fairness and accuracy simultaneously. The results table and figures demonstrate that our proposed method outperforms CAF and other methods.

\end{document}